\documentclass[lettersize,journal]{IEEEtran}
\usepackage{amsmath,amsfonts}
\usepackage{array}
\usepackage[caption=false,font=normalsize,labelfont=sf,textfont=sf]{subfig}
\usepackage[font=normalsize]{caption}
\usepackage{textcomp}
\usepackage{stfloats}
\usepackage{url}
\usepackage{verbatim}
\usepackage{graphicx}
\usepackage{cite}
\usepackage{soul}
\usepackage{enumitem}
\usepackage[english]{babel}
\usepackage{algorithm}  
\usepackage{algorithmicx}  
\usepackage{algpseudocode}
\usepackage{amssymb}
\usepackage [pdftex]{hyperref}
\usepackage[utf8]{inputenc}
\usepackage[T1]{fontenc}    
\usepackage{xcolor}    
\usepackage{url}       
\usepackage{booktabs}      
\usepackage{amsfonts}      
\usepackage{nicefrac}      
\usepackage{microtype}      
\usepackage{amsfonts}
\usepackage{amsmath}
\usepackage{multirow}
\usepackage{booktabs}
\usepackage{graphics}
\usepackage{epsfig}
\usepackage{titlesec}
\titlespacing\subsection{0pt}{12pt plus 4pt minus 2pt}{0pt plus 2pt minus 2pt}
\usepackage{color} 
\newcommand{\RNum}[1]{\uppercase\expandafter{\romannumeral #1\relax}}

\usepackage{hyperref}
\usepackage{float}
\usepackage{subfig}

\newcommand{\mb}{\mathbf}
\newcommand{\mt}{\mathcal}

\begin{document}
\setstcolor{red}
\title{Explaining Dynamic Graph Neural Networks \\via Relevance Back-propagation}

\author{Jiaxuan Xie, Yezi Liu, Yanning Shen$^\ast$
\IEEEcompsocitemizethanks{$^\ast$ Corresponding Author (yannings@uci.edu).}
\IEEEcompsocitemizethanks{Part of this work is supported by the Hellman Fellowship and the Google Research Scholar Award. All authors are with the Department of Electrical Engineering and Computer Science, University of California Irvine, Irvine, CA 92623 USA (e-mails: Jiaxux2@uci.edu, yezil3@uci.edu, yannings@uci.edu).}
}
\maketitle
\begin{abstract}
Graph Neural Networks (GNNs) have shown remarkable effectiveness in capturing abundant information in graph-structured data.
However, the black-box nature of GNNs hinders users from understanding and trusting the models, thus leading to difficulties in their applications.
While recent years witness the prosperity of the studies on explaining GNNs, most of them focus on static graphs, leaving the explanation of dynamic GNNs nearly unexplored. 
It is challenging to explain dynamic GNNs, due to their unique characteristic of time-varying graph structures.
Directly using existing models designed for static graphs on dynamic graphs is not feasible because they ignore temporal dependencies among the snapshots.
In this work, we propose DGExplainer to provide reliable explanation on dynamic GNNs. DGExplainer redistributes the output activation score of a dynamic GNN to the relevances of the neurons of its previous layer, which iterates until the relevance scores of the input neuron are obtained.
We conduct quantitative and qualitative experiments on real-world datasets to demonstrate the effectiveness of the proposed framework for identifying important nodes for link prediction and node regression for dynamic GNNs.
\end{abstract}

\begin{IEEEkeywords}
Explainable AI, Dynamics, Graph Neural Networks, Layer-wise Relevance Propagation
\end{IEEEkeywords}
\section{Introduction}\label{sec:intro}
A graph is a ubiquitous data structure that models the pairwise interactions of entities. For example, in social, information, chemical, and biological domains, the data can be naturally modeled as graphs. Graph Neural Networks (GNNs) have emerged as state-of-the-art for tackling graph data ~\cite{kipf2016semi, hamilton2017inductive}. 
While many real-world graphs are dynamic and evolving over time, various GNN models are proposed to deal with those dynamic graph structures~\cite{sankar2018dynamic,ma2020streaming,pareja2020evolvegcn}.
Though GNNs make useful predictions, the black-box nature hinders users from understanding and trusting their predictions.

The studies on explainability for static GNNs have attracted great interest in recent years. Some methods are approximation-based, which use gradient or surrogate functions to approximate the target instance output from a local model~\cite{huang2020graphlime,baldassarre2019explainability,pope2019explainability}.
However, a good approximation does not guarantee a good explanation in the sense of fidelity, due to the fact that the surrogate which approximates the target model possibly uses distinct features~\cite{rudin2019stop}.
And some other methods are perturbation-based~\cite{ying2019gnnexplainer,luo2020parameterized,schlichtkrull2020interpreting} via masking features, thus will generate an artificial impact on the model prediction and may trigger the adversarial property of deep neural networks~\cite{chang2018explaining}.
Additionally, gradient-based methods rest on an additive assumption of feature values or the gradients to measure the importance of the input features towards the output.
In summary, existing methods suffer from the adversarial triggering issue, fidelity issue, as well as additive assumption.  

Though urgently needed, it is a challenging task to design such an algorithm without the above issues.
Dynamic graphs are represented as a sequence of graph snapshots from different timesteps, so how to model this temporal evolutionary pattern is a crucial problem.
To this end, we propose a framework DGExplainer (Dynamic Graph neural network Explainer) to provide a faithful explanation for dynamic GNNs. DGExplainer uses a backward propagation manner to compute the relevance of the output corresponding to each input feature in the dynamic GNN model. Similar to many recent backward-based methods \cite{zhang2018top,bach2015pixel, 2020Explainability,rebuffi2020there}, the basic idea is to compute a small subset of node features that are most important to the prediction as an explanation.
First, it decomposes the prediction of a dynamic GNN and computes the relevances in a time-related module by employing Layer-wise relevance propagation (LRP). 
Then it further computes the relevances of the input features by back-propagating the relevances of the graph-related modules in the graph at each timestamp. Finally, by aggregating the obtained relevances from the above two steps, we can get the final relevances of node features as the measurement of the importance of the prediction.
Here we list our main contributions:
\begin{itemize}
\item We formally define the problem of explainability of dynamic graph neural networks. As far as we know, it is one of the first few attempts to study this problem.
\item We propose a novel framework (DGExplainer) to generate explanations for dynamic GNNs, from the perspective of decomposition. DGExplainer can effectively calculate the relevance scores which represent the contributions of each component for a dynamic graph.
\item We demonstrate the effectiveness of the DGExplainer on six real-world datasets. The quantitative experiments indicate that the proposed method could provide faithful explanations. The qualitative experiments show that the proposed method can encode time dependency in relevance scores at different timesteps.
\end{itemize}
	
\section{Methodology}\label{sec:method}
In this section, we elaborate on the proposed methods in detail. First, we introduce the notations and problem definition. Then, we elucidate the general setting of dynamic graph neural network models. Finally, we develop the proposed method for explaining dynamic GNNs.
\begin{figure}[t]
	\centering
	\includegraphics[width=1.0\linewidth]{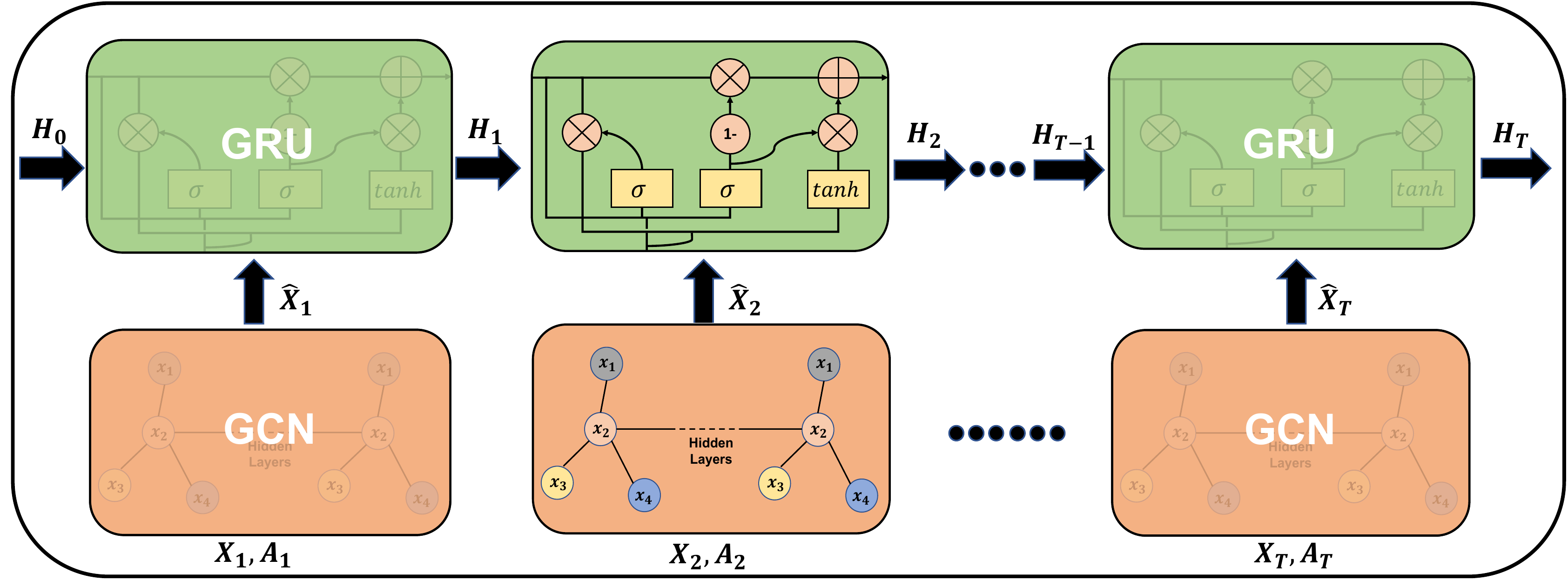}
	\caption{The network structure of the GCN-GRU model. Note that the GRUs and GCNs share the same parameters.
		$\{   \mathbf{H}_t\}_{t=0}^{T}$, $\{ \mathbf{X}_t\}_{t=1}^{ T}$, $\{\mathbf{\hat{X}}_t\}_{t=1}^{T}$, $\{\mathbf{A}_t\}_{t=1}^{T} $
		denote hidden features in GRU, node features, transformed features by GCN, adjacency matrices in different time slots, respectively.
		} 
	\label{fig:omega}
	\vspace{-2mm}
\end{figure}

\subsection{Problem Definition}
Given a dynamic graph as a sequence of snapshots $\mt{G} = \{ \mt{G}_t\}_{t=1}^{T}$, 
where $T$ is the length of the sequence. $\mt{G}_t = \{ \mt{V}_t, \mt{E}_t\}$ represents the graph at time $t$ and $\mt{V}_t$, $\mt{E}_t$ represents the node set, the edge set, respectively. The adjacency matrix at timestamp $t$ is represented as $\mb{A}_t\in \mathbb{R}^{N\times N}$, where $N=|\mt{V}_t|$ is the number of nodes. The feature matrix is denoted as $\mb{X}_t\in \mathbb{R}^{N\times D}$, where $D$ is the feature dimension, and $\mb{x}_t^i =  (\mb{X}_t^{(i,:)})^{\top} \in \mathbb{R}^{D}$ is the attribute vector for node $i$ at timestamp $t$, $\textit{i.e.}$ the $i$th row of $\mb{X}_t$. 
Without loss of generality, here we exploit $\mathbf{A}^{(i,j)}$ to denote the entry at $i$-th row, $j$-th column of adjacency matrix $\mathbf{A}$, and use $\mathbf{x}^{(i)}$ to denote the $i$-th entry of vector $\mathbf{x}$.
We leverage $R_{k}$ to denote the relevance of $k$, where $k$ can represent a node, an edge, a feature, etc. Also, we use $R_{k_1\leftarrow k_2}$ to denote the relevance of $k_1$ is distributed from $k_2$.
The goal of explaining dynamic GNNs is to find the subgraph in $\mt{G}$ that is the most important at timestamp $t$, given a dynamic GNN model $f(\mt{G})$. 
\subsection{Relevance-Propagation Based Explanation}
\label{LRP-sec}
Layer-wise relevance propagation (LRP) is first proposed in \cite{bach2015pixel}, which aims to compute the contribution (relevance) of each pixel for the prediction of an instance given the image classifier.
It assumes that the relevance is proportional to the weighted activation value this neuron produces, which follows the intuition that the larger the output activation, the more information this neuron carries, and the more contribution it has to the result.
The relevance score of the output neuron is defined as the prediction score. To compute the relevance of an individual neuron, the proposed framework operates by back-propagating the relevance scores from high-level neurons recursively to low-level neurons layer by layer, finally reach to the input neurons. The propagation rule can be summarized as follow:
\begin{align}
R_{k_1\leftarrow k_2}^{(l)} = \sum_{k_2} \frac{\mb{W}_{k_{1}k_{2}}a_{k_1}^{(l)}}{\epsilon+
\sum_k \mb{W}_{k k_{2}}a_{k}^{(l)}
}R_{k_2}^{(l+1)},
\label{LRP_propa}
\end{align}\\
where $\mb{W}_{k_{1}k_{2}}$ is the connection weight of neuron $k_{1}$ and $k_{2}$,
$R_{k_2}^{(l+1)}$ and $R_{k_1\leftarrow k_2}^{(l)}$ are relevances for the neuron $k_2$ at layer $l+1$ and the relevance of $k_1$ from $k_2$ at layer $l$, $a_{k_1}^{(l)}$ is the activation of neuron $k_1$ at layer $l$, and 
$\epsilon$ is a predefined stabilizer that prevent the denominator to be zero.
The main challenge for layer-wise relevance propagation is to design a proper propagation rule for the relevance redistribution of each layer. 
\subsection{Dynamic graph neural network model}
A dynamic graph neural network (DGNN) takes a dynamic graph as input and outputs a graph with updated topology, node, and/or edge information.
In this work, we exploit the GCN-GRU as our basic model. We summarize our model structure in Fig. \ref{fig:omega}.
Specifically, the graph convolution is defined as follows:
\begin{equation}
\mathbf{F}^{(l+1)}_t = \sigma(\mathbf{V}_t\mathbf{F}^{(l)}_t\mathbf{W}^{(l)}_t), \label{eq15}
\end{equation}
where $\mathbf{V}_t:=\tilde{\mathbf{D}}_t^{-\frac{1}{2}} \tilde{\mathbf{A}}_t \tilde{\mathbf{D}}_t^{-\frac{1}{2}}$ is the normalized adjacency matrix, and $\mb{D}_t$ is the degree matrix with $\mb{D}^{(i,i)}_t = \sum_j \mb{A}^{(i,j)}_t$, $\tilde{\mathbf{A}}_t = \mathbf{A}_t+{I_N}, \tilde{\mathbf{D}}_t = \mathbf{D}_t+\mathbf{I}_N$, $\mb{I}_N$ is an identity matrix with size $N$. $ \mathbf{F}^{(l)}_t$ is the output at the $l$ th layer, and $\mb{F}_t^{(0)}=\mb{X}_t$. Here we assume the GCN has $M$ layers, thus the final representation contains graph structural information is $\hat{\mb{X}}_t = \mb{F}_t^{(M)}$.  

Then the representation $\{\hat{\mb{X}}_t\}_{t=1}^{T}$ obtained from GCN, is fed into GRU. 
A GRU cell is a basic unit in GRU, where the input is $\mathbf{\hat{x}}_t={(\hat{\mb{X}}_t^{(i,:)})}^{\top}$, the hidden state is $\mathbf{{h}}_{t}=(\mb{H}_{t}^{(i,:)})^{\top}$.
Then the update rule of a GRU cell is shown below:
\begin{subequations}
	\begin{align}
	\mathbf{r}&=\sigma\left(\mathbf{W}_{i r} \mathbf{\hat{x}}_t+\mathbf{b}_{i r}+\mathbf{W}_{h r} \mathbf{{h}}_{t-1}+\mathbf{b}_{h r}\right) \label{eq3}\\ 
	\mathbf{z}&=\sigma\left(\mathbf{W}_{i z} \mathbf{\hat{x}}_t+\mathbf{b}_{i z}+\mathbf{W}_{h z} \mathbf{{h}}_{t-1}+\mathbf{b}_{h z}\right) \label{eq4}\\
	\mathbf{n}&=\tanh \left(\mathbf{W}_{i n} \mathbf{\hat{x}}_t+\mathbf{b}_{i n}+\mathbf{r} \odot\left(\mathbf{W}_{h n} \mathbf{{h}}_{t-1}+\mathbf{b}_{h n}\right)\right) \label{eq5}\\
	\mathbf{{h}}_{t}&=(1-\mathbf{z}) \odot \mathbf{{h}}_{t-1}+\mathbf{z} \odot \mathbf{n}, \label{eq6}
	\end{align}
	\label{eq2all}
\end{subequations}
where $\mathbf{W}_{ir}, \mathbf{W}_{hr}, \mathbf{W}_{hz}, \mathbf{W}_{in}, \mathbf{W}_{hn}$, $\mathbf{b}_{ir}, \mathbf{b}_{hr},  \mathbf{b}_{hz},  \mathbf{b}_{in},  \mathbf{b}_{hn}$ are learnable parameters in GRU, $\sigma(\cdot)$ denotes an activation function, and $\odot$ stands for an element-wise product operation. 
\begin{figure}[t]
	\centering
	\includegraphics[width=1.0\linewidth]{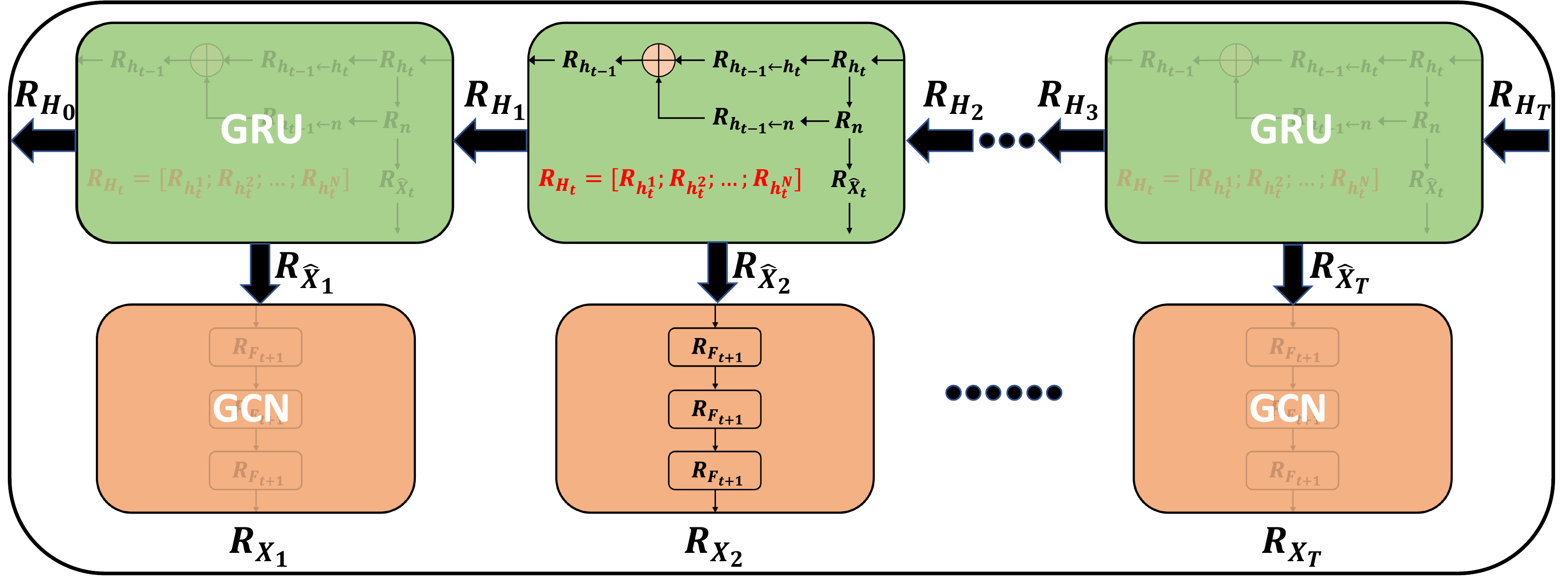}
	\caption{Illustration of the DGExplainer for calculating relevances in a backward manner. The relevance of the feature is calculated by back-propagating the final output $R_{\mb{h}_T}$ first through GRU and then via GCN.}
	\label{fig:r}
	\vspace{-3mm}
\end{figure}
\section{Explanation for dynamic GNNs}
Here, we introduce how to back-propagate the relevance through the GCN-GRU model for dynamic graphs. Fig. \ref{fig:r} demonstrates the layer-wise back-propagation process for calculating the relevance of the input features.
Given an input dynamic graph $\mt{G}=\{ \mt{G}_t\}_{t=1}^{T}$ and a dynamic GNN model $f(\mt{\cdot})$, for a graph snapshot $\mt{G}_t$, LRP aims to find out, which nodes contribute to what extent to a (positive or negative) classification or regression result by redistributing the prediction to the relevance of each neuron, such that the relevance of the neuron at the last layer is defined as $R_{\mb{h}_t}:=f(\mt{G}_t)$, where the task can be a classification (assuming the results from the classifier are thresholded at zero) or regression. For a single node, our method is a mapping: $\mathbb{R}^{N}\rightarrow\mathbb{R}^{1}$, such that the relevance denotes the extent of the contribution to the result.
\subsection{Compute the relevances in GRU}
Given $R_{\mb{h}_T}$ at $t=T$, the goal is to compute $R_{\mb{h}_{t-1}}$ and $R_{\hat{\mb{x}}_{t-1}}$ from $R_{\mb{h}_{t}}$. 
As described in Sec.~\ref{LRP-sec}, relevance back-propagation redistributes the activation of a descendant neuron to its predecessor neurons. And the relevance is proportional to the weighted activation value. 
Eqn.~\ref{eq6} shows the relation of different components in the last step of GRU. Note that, neurons $\mb{r}$ and $\mb{z}$ only have a message from neuron $\mb{h}_{t-1}$, as shown in Eqn.~\ref{eq3} and Eqn.~\ref{eq4}, so their message (contribution) to $\mb{h}_{t}$ could be merged into the message from neuron $\mb{h}_{t-1}$ and their relevance scores can be regarded as constants, so we do not have to consider their relevances.
Thus, based on Eqn.~\ref{eq6}, the relevance of components $\mb{h}_t$, $\mb{n}$, $\mb{h}_{t-1}$ satisfy:
\begin{equation}
R_{\mb{h}_t} =  R_{\mb{h}_{t-1}\leftarrow \mb{h}_t}+R_{\mb{n}\leftarrow \mb{h}_t}. \label{eqa8}
	\vspace{-2mm}
\end{equation}
Notice that $\mb{h}_{t-1}$ is used to compute $\mb{n}$ in Eqn. (\ref{eq5}) and $\mb{h}_t$ in Eqn. (\ref{eq6}), which reveals that $R_{\mb{h}_{t-1}}$ have two sources, 
$\mb{n}$ and  $\mb{h}_t$.
Therefore, we can define $R_{\mb{h}_t}$ based on $R_{\mb{h}_{t-1}\leftarrow\mb{n}}$ and $R_{\mb{h}_{t-1}\leftarrow \mb{h}_t}$ as:
\begin{equation}
R_{\mb{h}_{t-1}} =  R_{\mb{h}_{t-1}\leftarrow \mb{h}_t}+R_\mb{\mb{h}_{t-1}\leftarrow n}. \label{eqa88}
\end{equation}
It can be concluded that, if we can derive $R_{\mb{h}_{t-1}\leftarrow \mb{h}_t}$ and $R_\mb{\mb{h}_{t-1}\leftarrow n}$, we will be able to get $R_{\mb{h}_t}$. Thus, we formulate this problem into three steps, computing $R_{\mb{h}_{t-1}\leftarrow \mb{h}_t}$, $R_\mb{\mb{h}_{t-1}\leftarrow n}$, and $R_{\mb{h}_{t-1}}$, as below:

\noindent \textbf{(a) Computing $R_{\mb{h}_{t-1}\leftarrow \mb{h}_t}$:} First, from Eqn.~\ref{eq6}, we can obtain:

\begin{equation}
\frac{R_{\mb{n}\leftarrow \mb{h}_t}}{R_{\mb{h}_{t-1}\leftarrow \mb{h}_t}} =
\frac{\mathbf{z} \odot \mathbf{n}}{(1-\mathbf{z}) \odot \mathbf{{h}}_{t-1}} \label{eqa9}.
\end{equation}

Solving for \eqref{eqa8} and \eqref{eqa9} obtains:
\begin{align}
R_{\mb{n}\leftarrow \mb{h}_t} &= \frac{ \mathbf{z} \odot \mathbf{n}}{\mathbf{h}_t+\epsilon}\odot R_{\mathbf{h}_t}=R_{\mb{n}}, \label{al4}\\
R_{\mb{h}_{t-1}\leftarrow \mb{h}_t} &= \frac{(1-\mathbf{z})  \odot \mathbf{{h}}_{t-1}}{\mathbf{h}_t+\epsilon} \odot R_{\mathbf{h}_t}, \label{al5}
\end{align}

where $\epsilon>0$ is a constant introduced to keep the denominator non-zero.

Notice that the only ancestor neuron of $\mb{n}$ is $\mb{h}_t$, so here $R_{\mb{n}\leftarrow \mb{h}_t}$ is actually $R_{\mb{n}}$, so in the following left of section, we use $R_{\mb{n}}$ for simplicity.

\noindent \textbf{(b) Computing $R_\mb{\mb{h}_{t-1}\leftarrow n}$:}
Denote different components in Eqn. (\ref{eq5}) as:
\begin{subequations}
	\begin{align}
\mb{n}_1 :&= \mathbf{W}_{i n} {\mathbf{\hat{x}}_t}, \label{eq11}\\
\mb{n}_2 :&=\mathbf{r}\odot\left(\mathbf{W}_{h n} \mathbf{{h}}_{t-1}\right)=\mb{W}_{r n}\mb{h}_{t-1}, \label{eq12}\\
\mb{b_n} :&= \mb{b}_{in}+\mb{r}\odot\mb{b}_{h n}.
	\end{align}
\end{subequations}

Then their relevance satisfies:
\begin{align}
R_{\mb{n}} &= R_{\mb{n}_1}+R_{\mb{n}_2}+R_{\mb{b}_n},\\
R_{\mb{n}_1}&:{R_{\mb{n}_2}:R_{\mb{b}_{n}}} = \mb{n}_1:\mb{n}_2:\mb{b}_{n},
\end{align}

hence $R_{\mathbf{n}_1}$ and $R_{ \mathbf{n}_2}$ can be obtained as:
\begin{align}
R_{ \mb{n}_1}\!\!&=\!\!\frac{\mathbf{W}_{in}  \mathbf{\hat{x}}_t}{\epsilon+(\mathbf{W}_{in} \mathbf{\hat{x}}_t+\mathbf{b}_{i n}+\mathbf{r} \odot\left(\mathbf{W}_{h n} \mathbf{h}_{t-1}+\mathbf{b}_{h n}\right))}\!\odot\! R_{\mathbf{n}},\label{al11}\\
R_{\mb{n}_2}\!\!&=\!\!\frac{\mathbf{r}\odot (\mathbf{W}_{hn}\mathbf{h}_{t-1})}{\epsilon+(\mathbf{W}_{in} \mathbf{\hat{x}}_t+\mathbf{b}_{i n}+\mathbf{r}\odot\left(\mathbf{W}_{h n} \mathbf{h}_{t-1}+\mathbf{b}_{h n}\right))}\!\odot\! R_{\mathbf{n}}.
\label{al12}
\end{align}

Let $\mb{n}_1^{(k)}$ denote the $k$-th entry of $\mb{n}_1$, according to (\ref{eq11}), we have $\mb{n}_1^{(k)} = \sum_j \mb{W}_{in}^{(k,j)}\mb{\hat{x}}_t^{(j)}$.
The relevance $R_{\mb{n}_1}$ is redistributed in proportion to the contribution for $\mb{n}_1$ and hence $R_{\hat{\mb{x}}_t}$, {which equals to $R_{\mb{\hat{x}}_{t}\leftarrow {\mb{n}_1}}$ because $\mb{n_1}$ is the only source of the relevance to $\hat{\mb{x}}_t$} by using LRP-$\epsilon$ rule \cite{bach2015pixel}: 
\begin{align}
R_{\mb{\hat{x}}_{t}\leftarrow {\mb{n}_1}} = \sum_k \frac{\mb{W}_{in}^{(k,j)}\mb{\hat{x}}_t^{(j)}}{\epsilon+ \sum_i \mb{W}_{in}^{(k,i)}\mb{\hat{x}}_t^{(i)}}R_{\mb{n}_1}^{(k)}. \label{al13} 
\end{align}
Since $\mb{h}_{t-1}$ only influences $\mb{n}_2$ among the three parts of $\mb{n}$, we obtain $R_{\mb{h}_{t-1}\leftarrow \mb{n}}$ using $\epsilon$-rule for Eqn.~\ref{eq12}:
\begin{align}
R_{\mb{h}_{t-1}\leftarrow \mb{n}}^{(j)} = \sum_k \frac{\mb{W}_{rn}^{(k,j)}\mb{h}_{t-1}^{(j)}}{\epsilon+ \sum_i \mb{W}_{rn}^{(k,i)}\mb{h}_{t-1}^{(i)} 
}R_{\mb{n}_2}^{(k)}. \label{al14}
\end{align}
\textbf{(c) Compute $R_{\mb{h}_{t-1}}$:} Upon obtaining $R_\mb{n\leftarrow \mb{h}_t}$, and $R_{\mb{h}_{t-1}\leftarrow \mb{h}_t}$ in {\ref{al4} and \ref{al5}},
based on Eqn.~\ref{eqa88},  $R_{\mb{h}_{t-1}}$ can be computed by adding Eqn.~\ref{al5} and  Eqn.~\ref{al14} together:
\begin{equation}
R_{\mb{h}_{t-1}} = R_{\mb{h}_{t-1}\leftarrow \mb{h}_t}+\sum_j R_{\mb{h}_{t-1}\leftarrow \mb{n}}^{(j)}. \label{final}
\end{equation}
Notice that $R_{\hat{\mb{x}}_t}$  is the relevance of  a sample $\hat{\mb{x}}_t$, which is a row in $\hat{\mb{X}}_t$. By computing $\{R_{\hat{\mb{x}}_t^i}\}_{i=1}^{N}$, 
we can get $R_{\hat{\mb{X}}_{t}} = [R_{\hat{\mb{x}}_{t}^1};R_{\hat{\mb{x}}_{t}^2};\ldots;R_{\hat{\mb{x}}_{t}^N}]$.
\subsection{Back-propagate the relevances in GCN}
Then we backtrack in the GCN to get $R_{\mb{X}_t}$ from $R_{\hat{\mb{X}}_t}$.
Note that the $R_{\mathbf{\hat{X}}_t}$ is the relevance of the output $\mathbf{\hat{X}}$ of the GCN at the time step $t$ and $R_{\hat{\mb{X}}_t} = R_{\mb{F}_t^{(M)}}$.
We can rewrite Eqn. (\ref{eq15}) as:
\begin{align}
\mathbf{F}_t^{(l+1)} &= \sigma(\mathbf{P}_t^{(l)}\mathbf{W}_t^{(l)}), \label{eq16}\\   \mathbf{P}_t^{(l)} :&= \mathbf{V}_t\mathbf{F}_t^{(l)}. \label{17}
\end{align}
Let  
$(\mathbf{F}_{t}^{(l+1)})^{(k,:)}$, 
$(\mathbf{P}_{t}^{(l)})^{(k,:)}$, 
$(\mathbf{P}_{t}^{(l)})^{(:,k)}$, 
$(\mathbf{F}_{t}^{(l)})^{(:,k)}$ denote  the $k$-th row of $\mathbf{F}_t^{(l+1)}$, the $k$-th row of $\mathbf{P}_t^{(l)}$, the $k$-th column of $\mathbf{P}_t^{(l)}$, the $k$-th column of $\mathbf{F}_t^{(l)}$, respectively. We have
\begin{align}
(\mathbf{F}_{t}^{(l+1)})^{(k,:)} &= \sigma((\mathbf{P}_{t}^{(l)})^{(k,:)}\mathbf{W}_t^{(l)}), \label{eq18}\\
(\mathbf{P}_{t}^{(l)})^{(:,k)} &:= \mathbf{V}_t(\mathbf{F}_{t}^{(l)})^{(:,k)}. \label{19}
\end{align}
Leveraging the $\epsilon$ rule, we assign the relevance by:
\begin{align}
R_{(\mathbf{P}_{t}^{(l)})^{(k,j)}}\!&=\!\sum_ b \frac{(\mathbf{P}_{t}^{(l)})^{(k,j)} (\mathbf{W}_{t}^{(l)})^{(j,b)} }  {\epsilon+ \sum_i (\mathbf{P}_{t}^{(l)})^{(k,i)} (\mathbf{W}_{t}^{(l)})^{(i,b)}
}
R_{(\mathbf{F}_{t}^{(l+1)})^{(k,b)}},\label{eq23}\\
R_{(\mathbf{F}_{t}^{(l)})^{(j,k)}}\!& =\!\sum_b\frac{ \mathbf{V}^{(b,j)} (\mathbf{F}_{t}^{(l)})^{(j,k)}     }{\epsilon + \sum_a\mathbf{V}_t^{(b,a)} (\mathbf{F}_{t}^{(l)})^{(a,k)}}R_{(\mathbf{P}_{t}^{(l)})^{(b,k)}},\label{eq24}
\end{align}
where $(\mathbf{W}_{t}^{(l)})^{(j,k)}$ represents the entry at $j$-th row and $k$-th column  of $\mathbf{W}_t^{(l)}$, and $\mathbf{V}_t^{(b,j)}$ denotes the entry at $b$-th row and $j$-th column of $ \mathbf{V}_t^{(k,j)}$.
$R_{\mb{F}_t^{(l)}}$ can henceforth be obtained from $R_{\mb{F}_t^{(l+1)}}$ via Eqn. (\ref{eq23}) and Eqn. (\ref{eq24}), and finally $R_{\mathbf{F}_t^{(0)}}$ can be obtained.
Notice that $\mathbf{F}_t^{(0)} = \mathbf{X}_{t}$, so we have $ R_{\mathbf{F}_t^{(0)}} =  R_{\mathbf{X}_{t}}$ and hence the backward process for obtaining relevance in the GCN is completed.
To further identify important nodes at a specific time-slot, we transform the relevances into absolute values sum up all the relevances along the {feature} dimension to get the relevance of a node at time $t$: $R_{\mb{x}^i_t}=\sum_{j=1}^D |(R_{\mb{x}^i_t})^{(j)}|$.
The entire algorithm is summarized in Algorithm \ref{code:fram:select}.
\renewcommand{\algorithmicrequire}{\textbf{Input:}} 
\renewcommand{\algorithmicensure}{\textbf{Output:}} 
\begin{algorithm}[!htb]
\small
	\caption{DGExplainer}
	\label{alg:Framwork}
	\begin{algorithmic}[1]
		\Require
		The input $\{\mb{X}_t\}_{t=1}^{T}$ and $\{ \mb{A}_t \}_{t = 1}^{T}$, the final relevance $\{R_{\mb{h}^j_{T}}\}_{j=1}^{N}$, the pre-trained model $f{(\cdot)}$.
		\Ensure
		The relevances $\{   R_{\mb{X}_t}  \}_{t=1}^{T}$
		\State $//$ Forward process:
		\For{each $t \in [1, T] $}
		\State Compute $ \hat{\mb{X}}_t$ via   $\mathbf{F}^{(l+1)}_t = \sigma(\mathbf{V}_t\mathbf{F}^{(l)}_t\mathbf{W}^{(l)}_t)$ with $\mb{F}_t^{(0)} = \mb{X}_t$, $\mb{F}_t^{(M)} = \hat{\mb{X}}_t$.
		\For{each $j \in [1, N]$}
		\State Compute the hidden state $\mb{h}_t$ for the $j$-th sample \State ${\hat{\mb{X}}}_t^{(j,:)}$ via Eqn. (\ref{eq2all}) with $\mb{h}_{t-1}$.
		\EndFor 
		\EndFor
		\State $//$ Backward process:
		\For{each $t = T, T-1, \ldots, 1 $}
		\For{each $j \in [1, N]$}
		\State Compute $R_{\mb{n}}, R_{\mb{n_1}}, R_{\mb{n_2}}$ via Eqn. (\ref{al4}, \ref{al11}, \ref{al12}), $R_{\mb{h}_{t-1}}$ via Eqn. (\ref{al5}, \ref{al14}, \ref{final}), and $R_{\hat{\mb{x}}_{t}}$ for the $j$-th sample ${\hat{\mb{X}}}_t^{(j,:)}$ via Eqn. (\ref{al13}) and hence obtain $R_{\hat{\mb{x}}_{t}^j}$.
		\EndFor 
		\State Stack $\{R_{\hat{\mb{x}}_{t}^j}\}_{j=1}^{N}$ to get $R_{\hat{\mb{X}}_{t}}$.
		\State Calculate $R_{\mb{X}_t}$ by iteratively applying Eqn. (\ref{eq23}-\ref{eq24}) with $R_{\hat{\mb{X}}_t} = R_{\mb{F}_t^{(M)}}$ and $R_{{\mb{X}}_t}$.
		\EndFor\\
		\Return $\{R_{\mathbf{X}_{t}}\}_{t=1}^{T}.$
	\end{algorithmic}
	\label{code:fram:select}
\end{algorithm}
\vspace{-5mm}
\section{Experiments}\label{sec:experiments}
In this section, we conduct extensive experiments on six real datasets in terms of three evaluation metrics to evaluate the performance of the proposed approach.
\vspace{-3mm}
\subsection{Datasets}
We evaluate our proposed method with other representative methods on real-world datasets. As described in Table~\ref{dataset}, the datasets are split into the training and test sets according to the temporal dimension. For each dynamic graph $\mt{G}$, given partially observed snapshots $\{ \mt{G}_1, \mt{G}_2, \ldots, \mt{G}_T\}$ with node attributes $\mt{X} = \{\mb{X}_1, \ldots, \mb{X}_T \}$ and adjacency matrices $\mt{A} = \{\mb{A}_1, \ldots, \mb{A}_T \}$, our goal is to predict the behavior of $\mt{G}_{T+1}$ for node attribution regression or dynamic link prediction. 
More details about the datasets can be found below.
\begin{table}[!htp]
			\setlength{\tabcolsep}{4pt}
	\centering
	\caption{The statistics of the datasets. }
	\begin{tabular}{lllllll}
		\toprule
		Dataset                         & RH   &{PeMS04}     &{PeMS08}  & Enron       &FB & COLAB      \\
		\midrule
		\# Nodes                    & 55,863           &307        &170   & 184       & 663   &315\\ 
		\# Edges  					& 858,490         &680        &340   & 266     &1068   &308  \\  
		\# Train/Test               & 122/34      &45/14    & 50/12   & 8/3  &  6/3   &7/3 \\ 
		\# Time Step  &   6        &   4     &    4          &   4     & 4   & 4\\ 
		\bottomrule
	\end{tabular}
	\label{dataset}
	\vspace{-2mm}
\end{table}
\begin{itemize}[leftmargin=*]
    \item \textbf{Reddit Hyperlink \cite{kumar2018community} (Reddit)}: it is a directed network extracted from posts that generate hyperlinks connecting one subreddit to another. 
    We use this dataset for link prediction.
    \item \textbf{PeMS04} and \textbf{PeMS08} \cite{attentionbased2019}: they are real-time traffic datasets, and are aggregated into every 5-minute interval from the raw data, so each detector contains 288 data points per day. The data are transformed by zero-mean normalization to let the average be 0. The task is node regression.
    \item \textbf{Enron}\cite{enron2004}, 	\textbf{FB} \cite{trivedi2018dyrep}, 	\textbf{COLAB} \cite{rahman2016link}:
    they are dynamic graphs constructed from email messages exchanged between employees, co-authorship relationships among authors, and the Facebook wall posts, correspondingly.
    We collect and tackle the three datasets following the previous work \cite{hajiramezanali2019variational}.
\end{itemize}
\begin{figure*}[b]
	\centering
\captionsetup[subfigure]{labelformat=empty,justification=centering}
\vspace{-3mm}
\noindent\begin{minipage}[b]{0.8\linewidth}
\centering
\hspace{-2mm}
		\subfloat[(a) Reddit]{\includegraphics[scale=0.31]{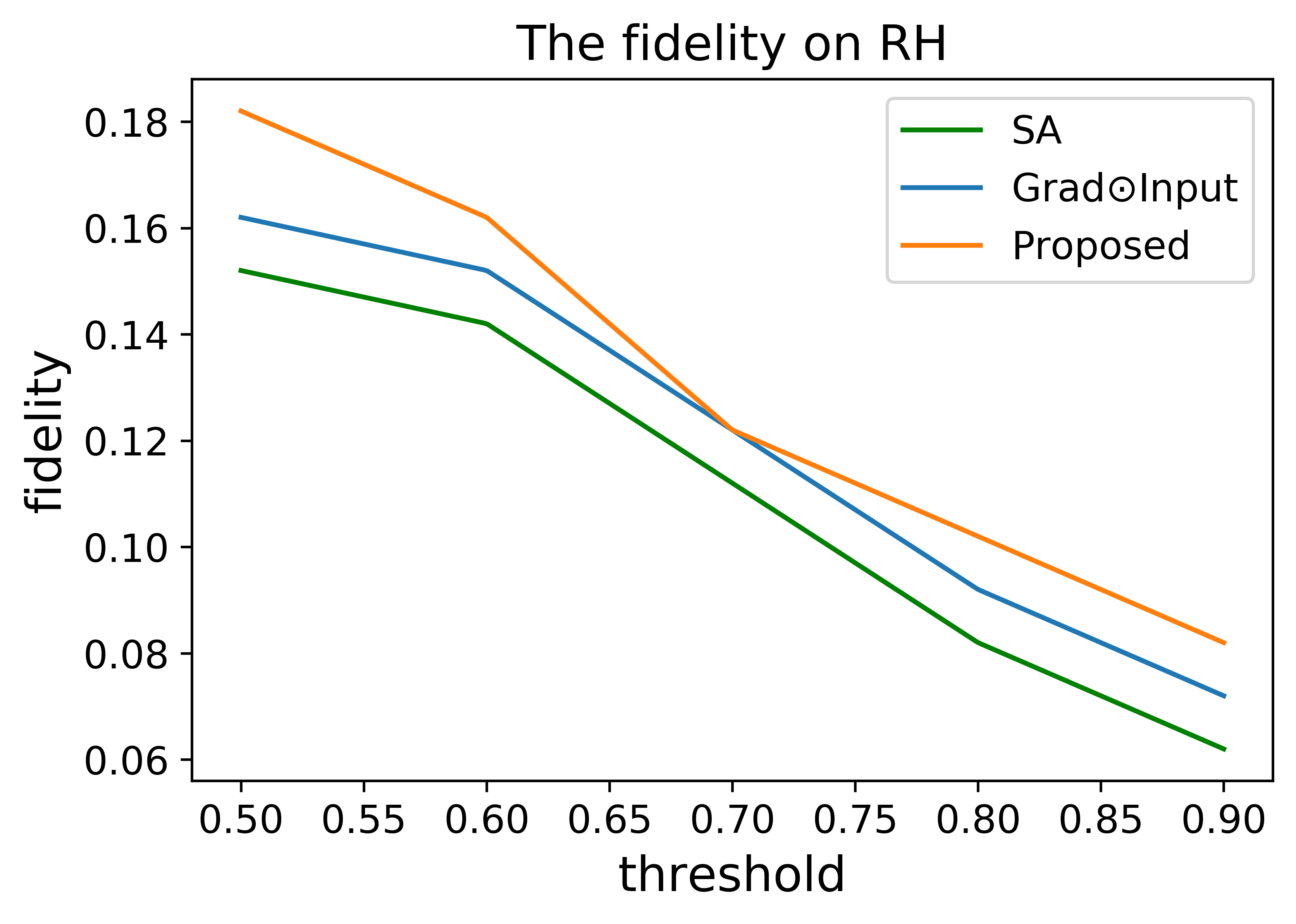}}
		\subfloat[(b) PeMS04]{\includegraphics[scale=0.31]{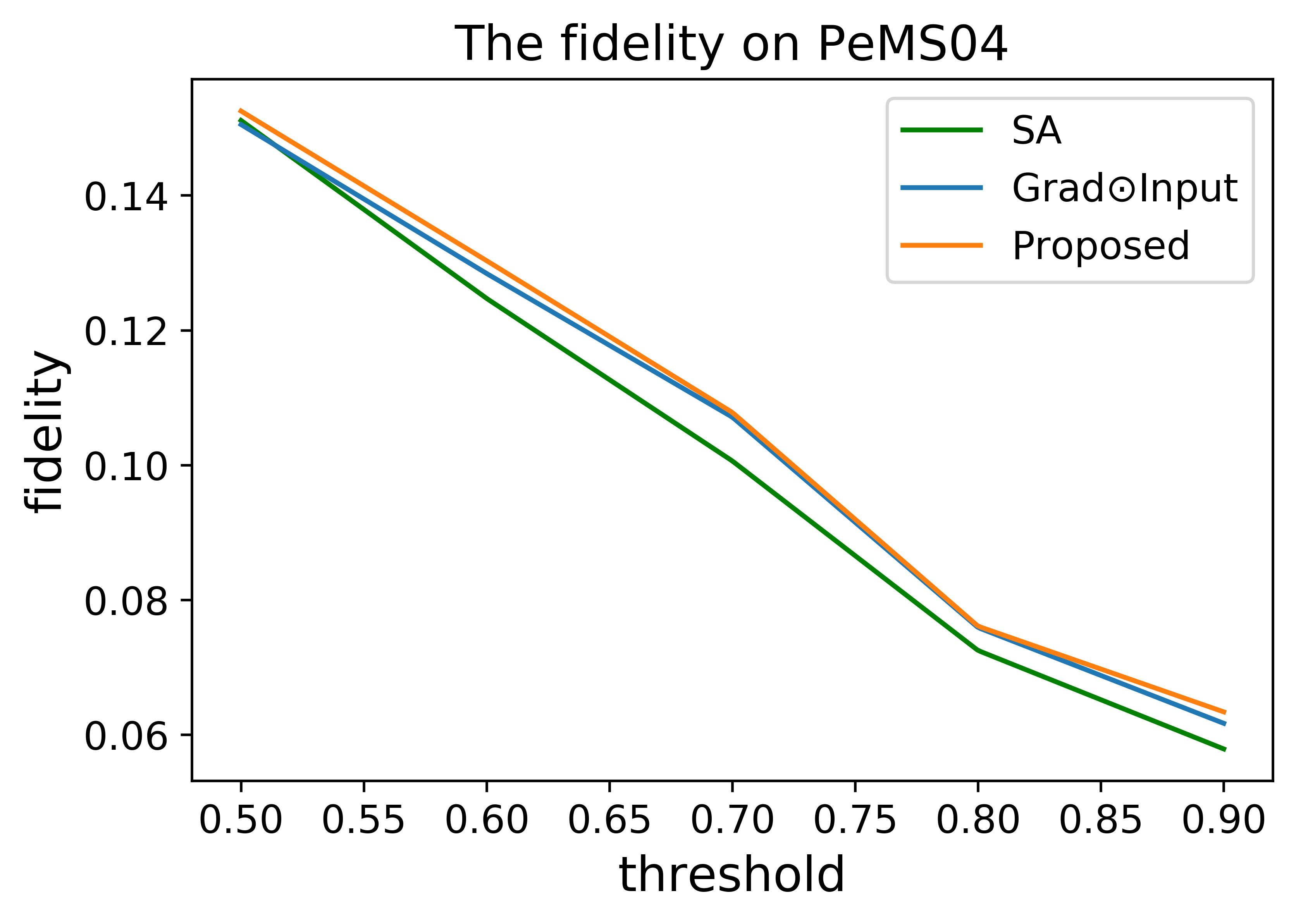}}
		\subfloat[(c) PeMS08]{\includegraphics[scale=0.31]{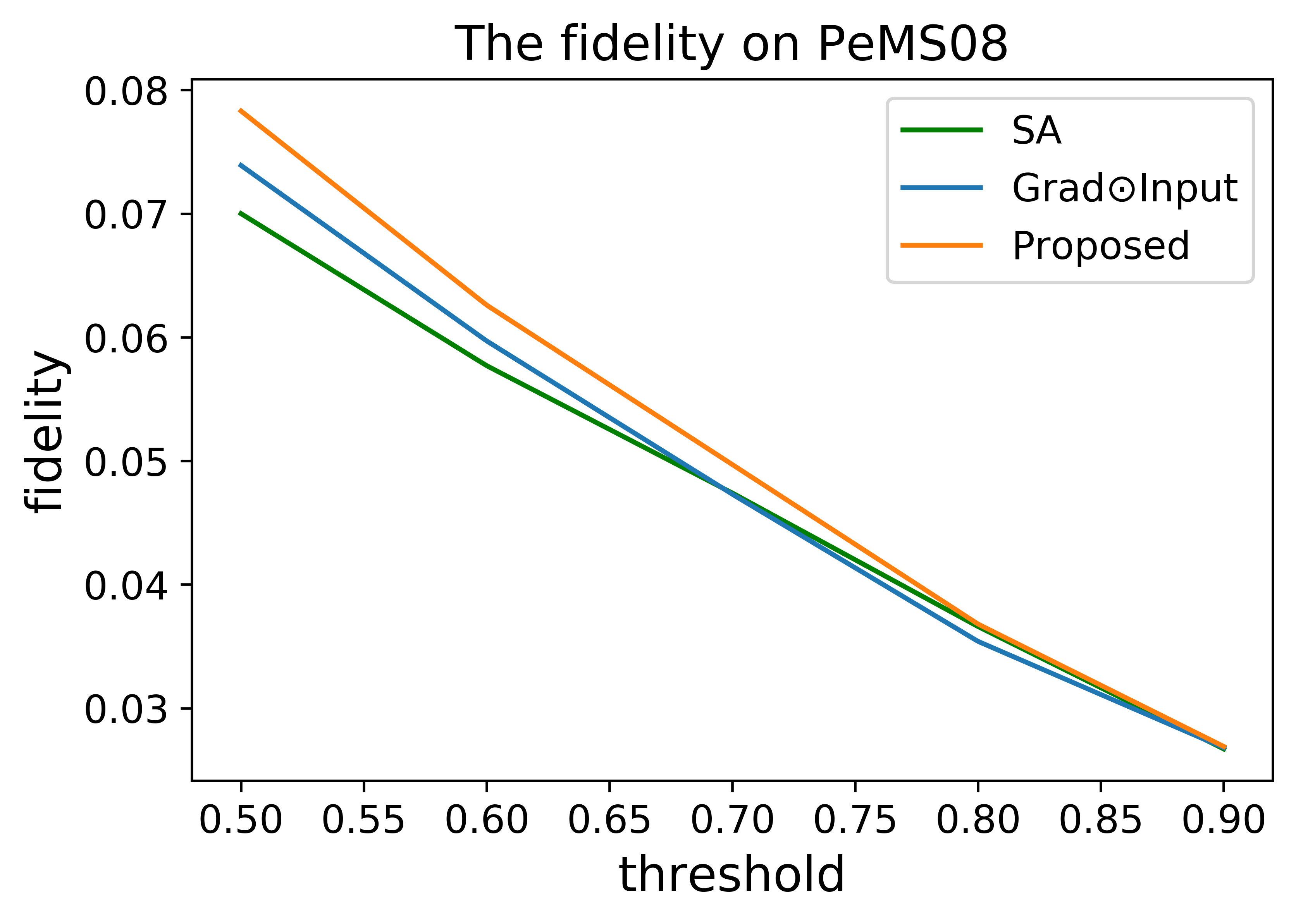}}
\end{minipage}
	\noindent\begin{minipage}[b]{0.8\linewidth}
	\centering
	\hspace{-2mm}
\subfloat[(d) Enron]{\includegraphics[scale=0.31]{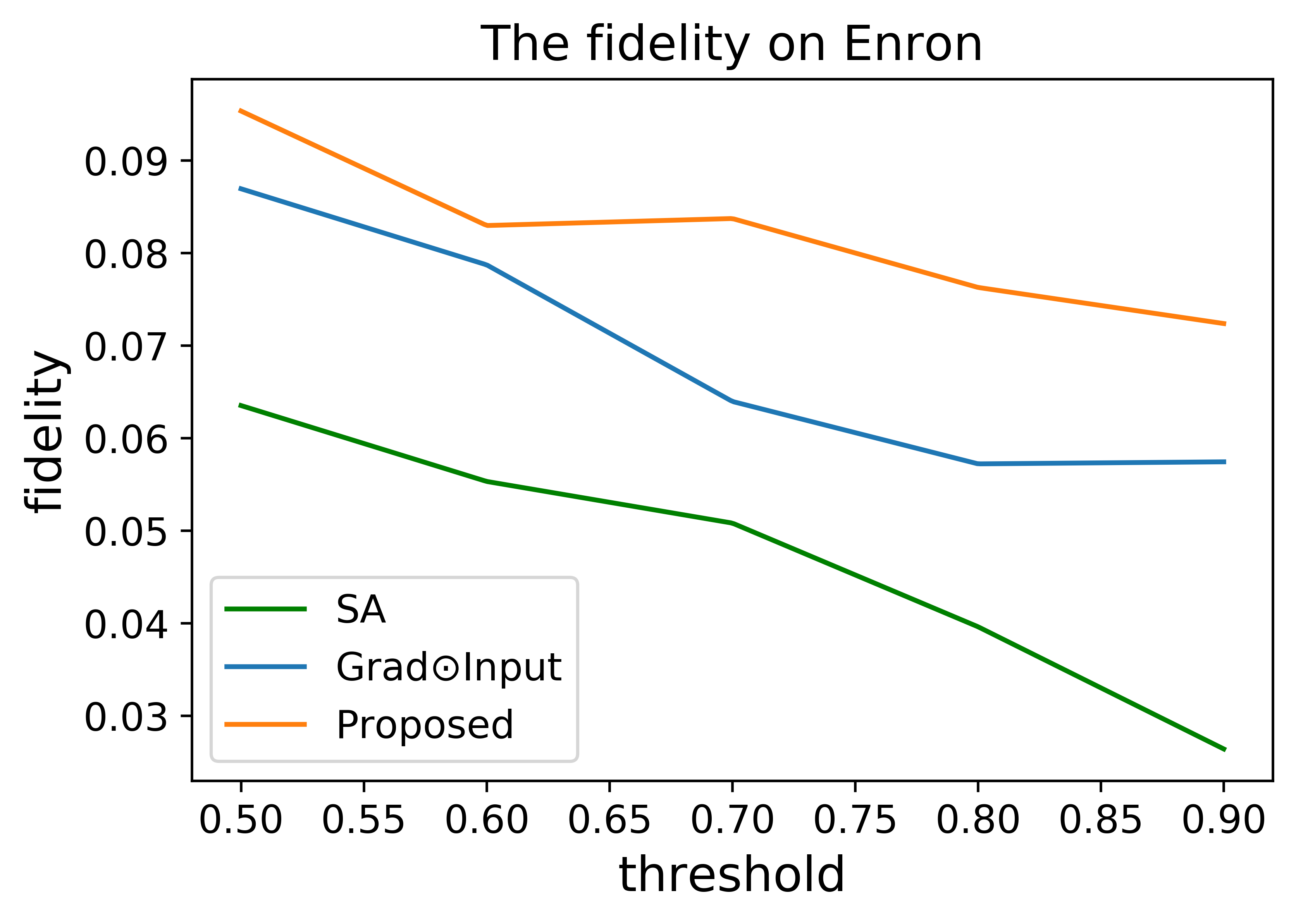}}
		\subfloat[(e) 
		FB]{\includegraphics[scale=0.31]{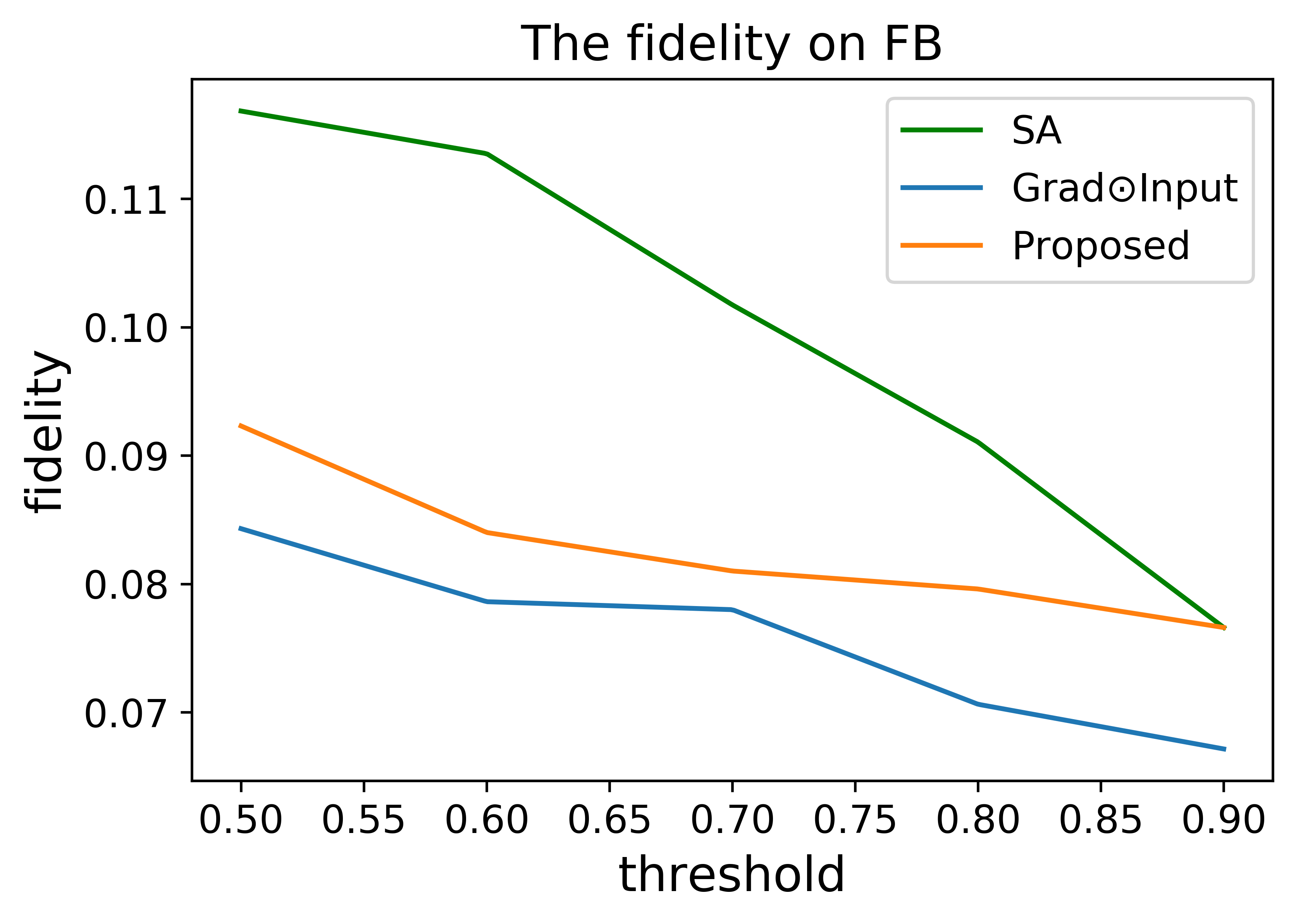}}
		\subfloat[(f) COLAB]{\includegraphics[scale=0.31]{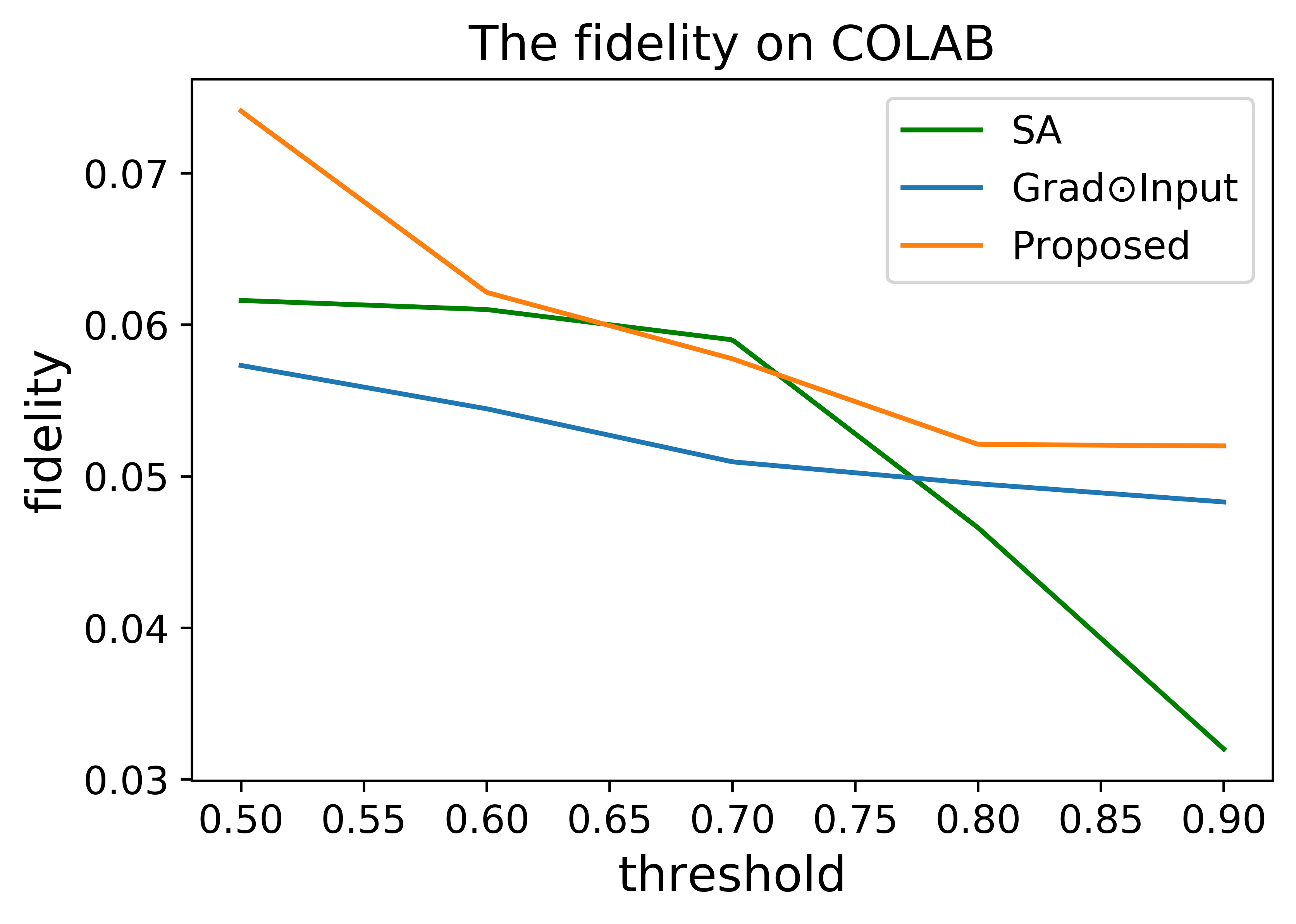}}
\end{minipage}
\centering
\caption{Comparison of fidelity measurement on different datasets.}
\label{fig:fidelity}
\end{figure*}
\vspace{-3mm}
\subsection{Tasks}
Following the setting in a previous work ~\cite{pareja2020evolvegcn}, we conduct experiments on two tasks, including link prediction and node classification. The output embedding of a node $u$ produced by the GCN-GRU model at time $t$ is represented by $\mb{h}_t^u$.

\noindent \textbf{Link prediction}:
For implementation, we concatenate the feature embeddings of $u$ and $v$ as $[(\mb{h}_T^u)^{\top}; (\mb{h}_T^v)^{\top}]^{\top}$
and apply an MLP to predict the link probability by optimizing the cross-entropy loss. Reddit Hyperlink, Enron, FB, and COLAB are used for experimentation for this task. We leverage Area Under Curve (AUC) as the evaluation metric.
\hfill\\
\noindent \textbf{Node regression}:
To get a prediction value for a node $u$ at time $t$, we exploit a softmax activation function for the last layer of GCN, outputting the provability vector $\mb{h}_t^u$. Two datasets are used for the experiments of this task, including PeMS04 and PeMS08. 
The evaluation metric is mean absolute error (MAE).
\subsection{Compared methods and evaluation indexes}
\subsubsection{Compared methods}
We compare the performance of our algorithm with two explainability methods:
\begin{itemize}[leftmargin=*] 
\item \textbf{Sensitivity Analysis (SA)
\cite{baldassarre2019explainability}}: It assumes that the norm of the gradient over input variables demonstrates their relative importance. Then the saliency map $M_{Grad}$ for the input $\mathbf{x}$ is computed by 
$
M_{Grad}=\left\|\frac{\partial \mathbf{y}^{c}}{\partial \mathbf{x}}\right\|,
$
where $\mathbf{y}^c$ is the score for class $c$ before a softmax layer.
\item \textbf{Grad$\odot$Input \cite{shrikumar2017learning}}: Grad$\odot$Input corresponds to the element-wise product of the input
with the gradient. The result $M_{GI}$ for Grad$\odot$Input is obtained by
$
M_{GI} =  \frac{\partial \mathbf{y}^c}{\partial \mathbf{x}} \odot \mathbf{x}. 
$
\end{itemize}
\begin{figure*}[t]
	\centering
\captionsetup[subfigure]{labelformat=empty,justification=centering}
\vspace{-3mm}
\noindent\begin{minipage}[b]{0.8\linewidth}
\centering
\hspace{-2mm}
		\subfloat[(a) RH]{\includegraphics[scale=0.31]{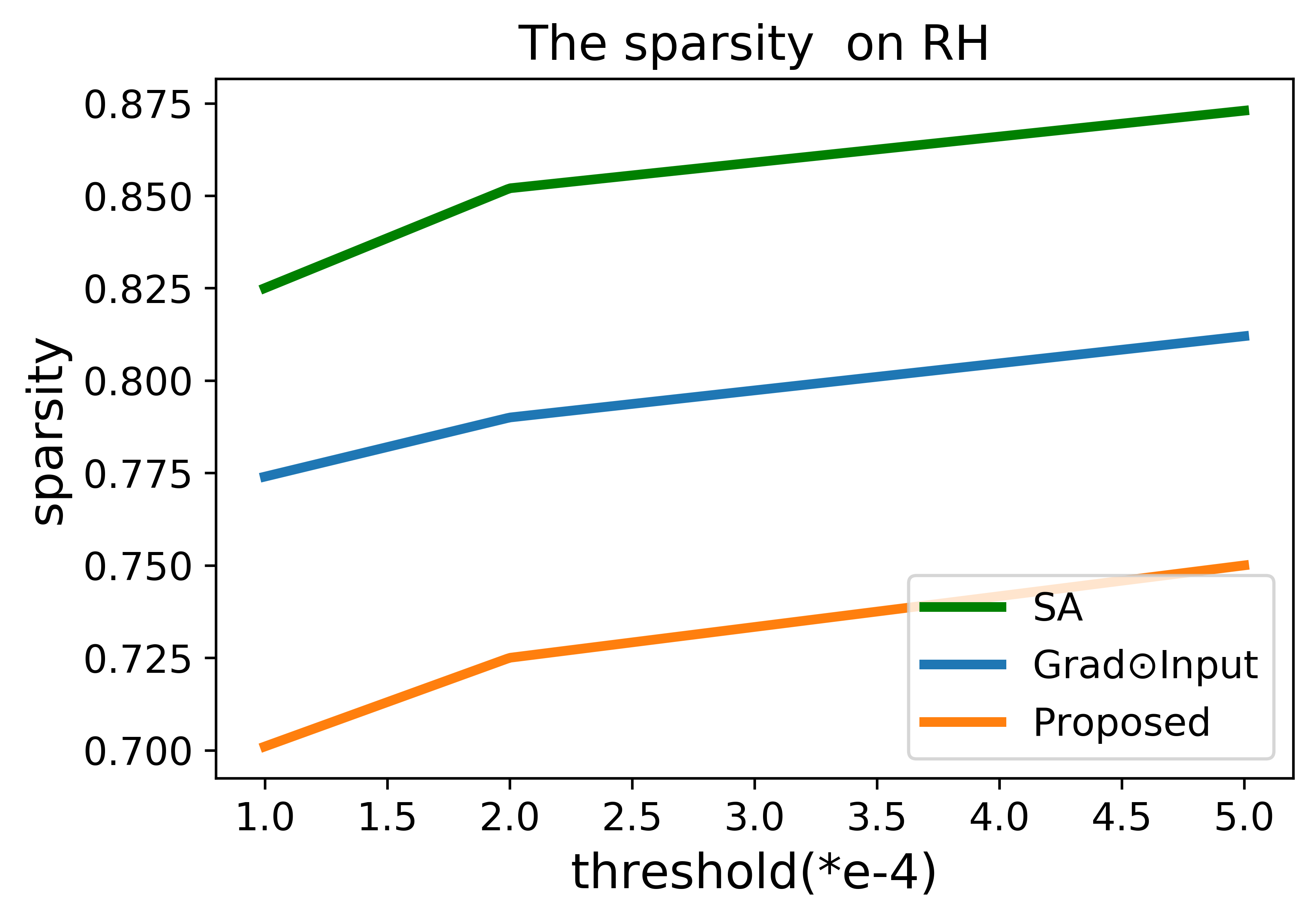}}
		\subfloat[(b)
		PeMS04]{\includegraphics[scale=0.31]{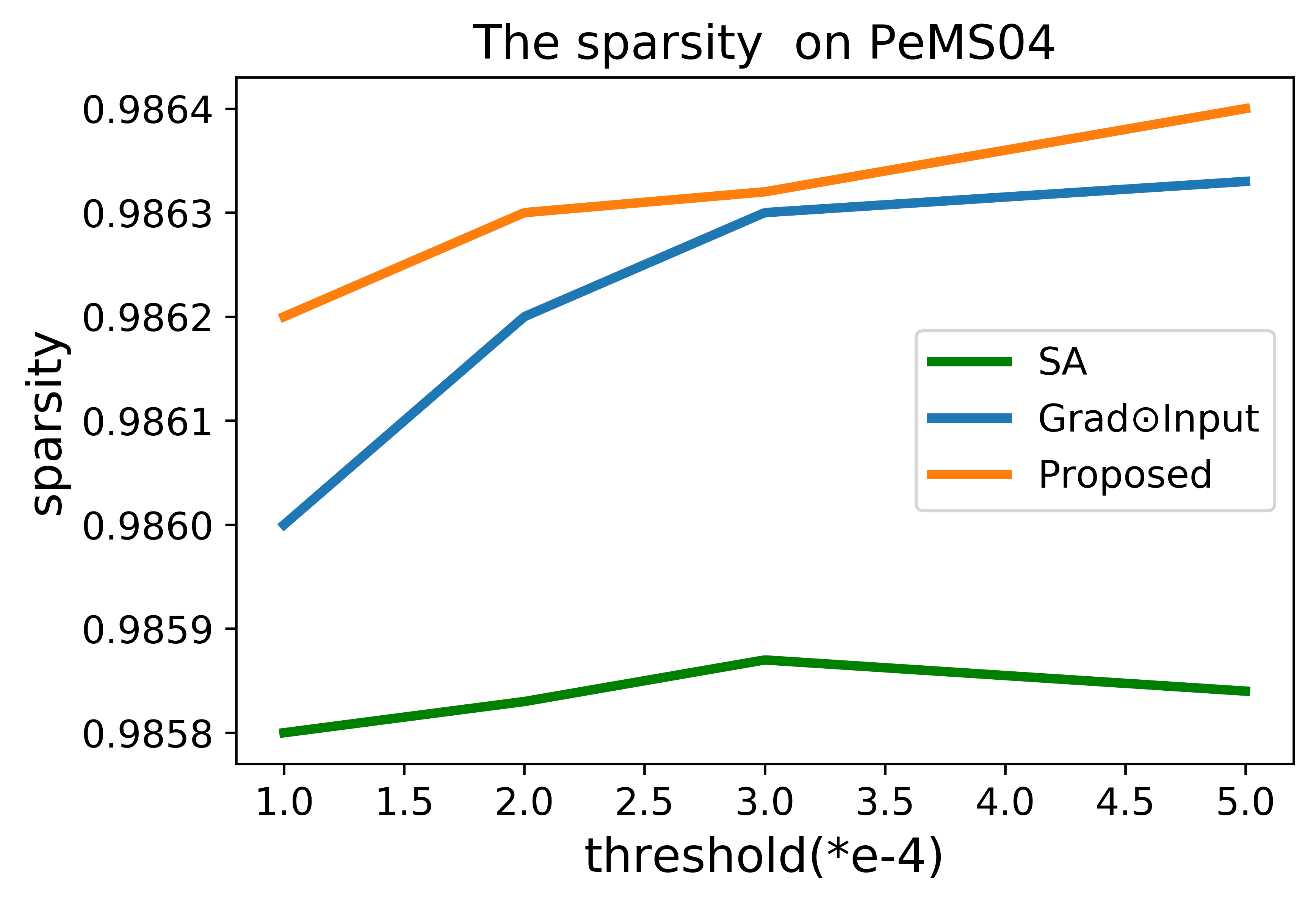}}
		\subfloat[(c) PeMS08]{\includegraphics[scale=0.31]{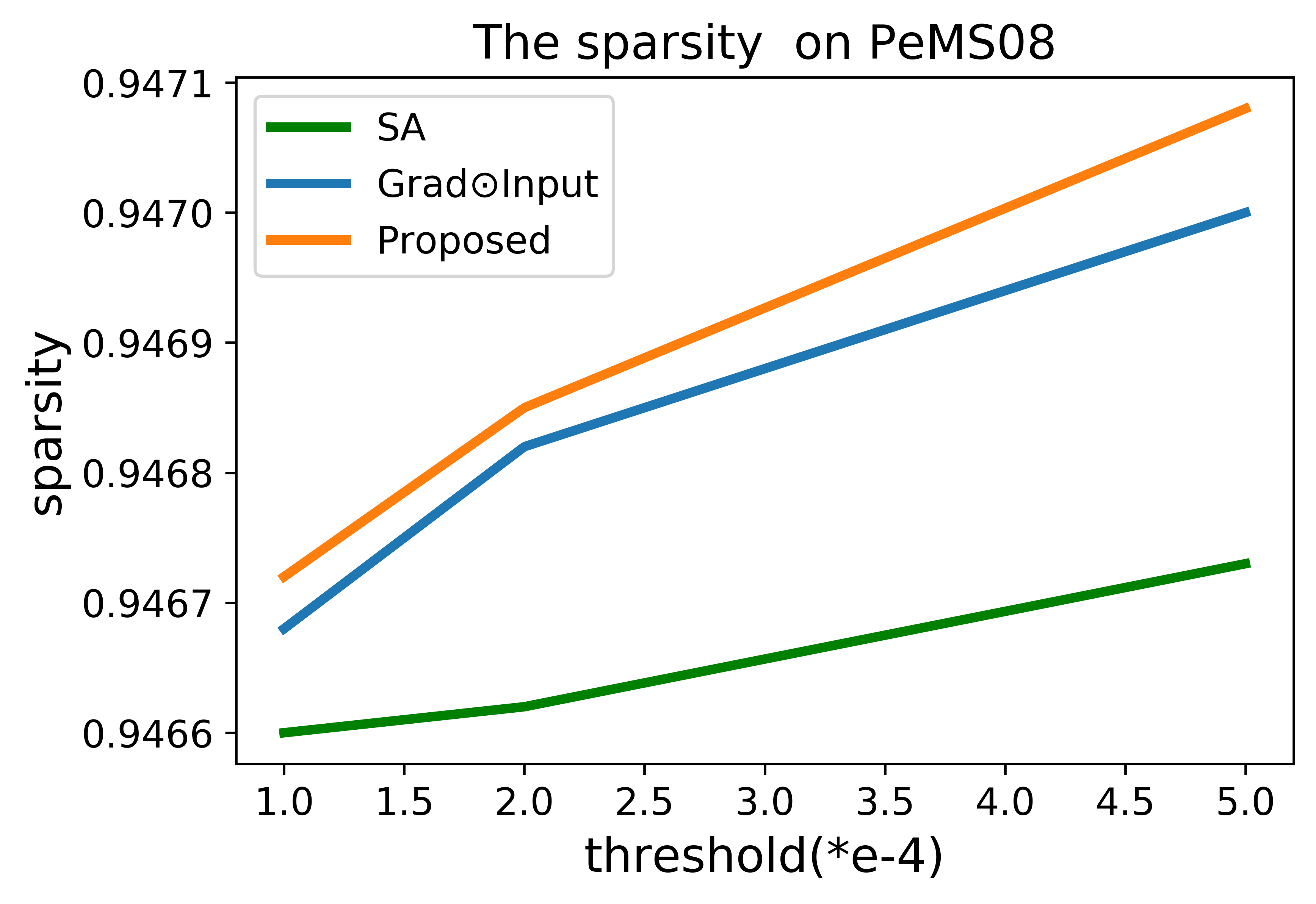}}
		\end{minipage}
		\noindent\begin{minipage}[b]{0.8\linewidth}
		\centering
			\hspace{-2mm}
		\subfloat[(d) Enron]{\includegraphics[scale=0.31]{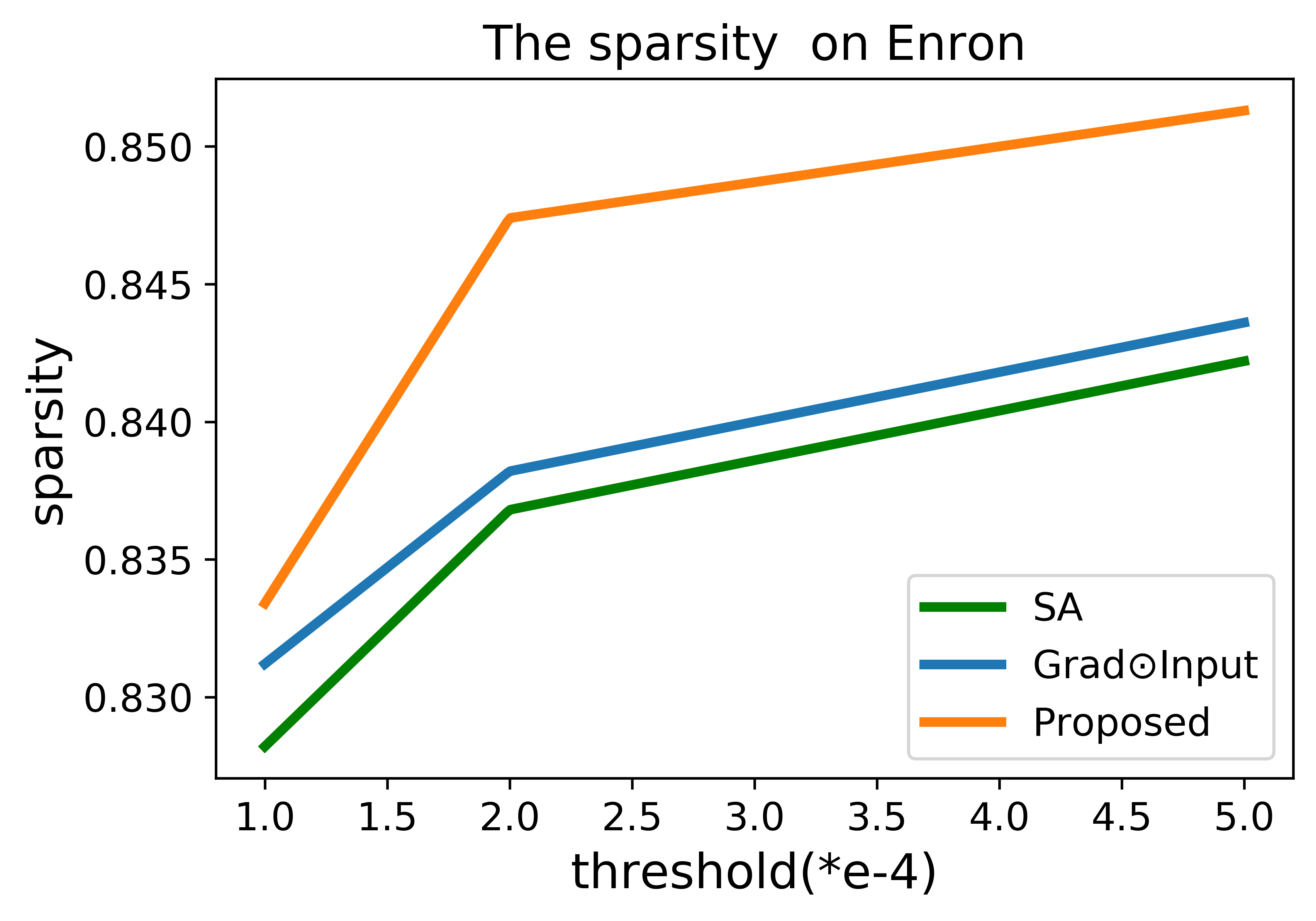}}
		\subfloat[(e) FB]{\includegraphics[scale=0.31]{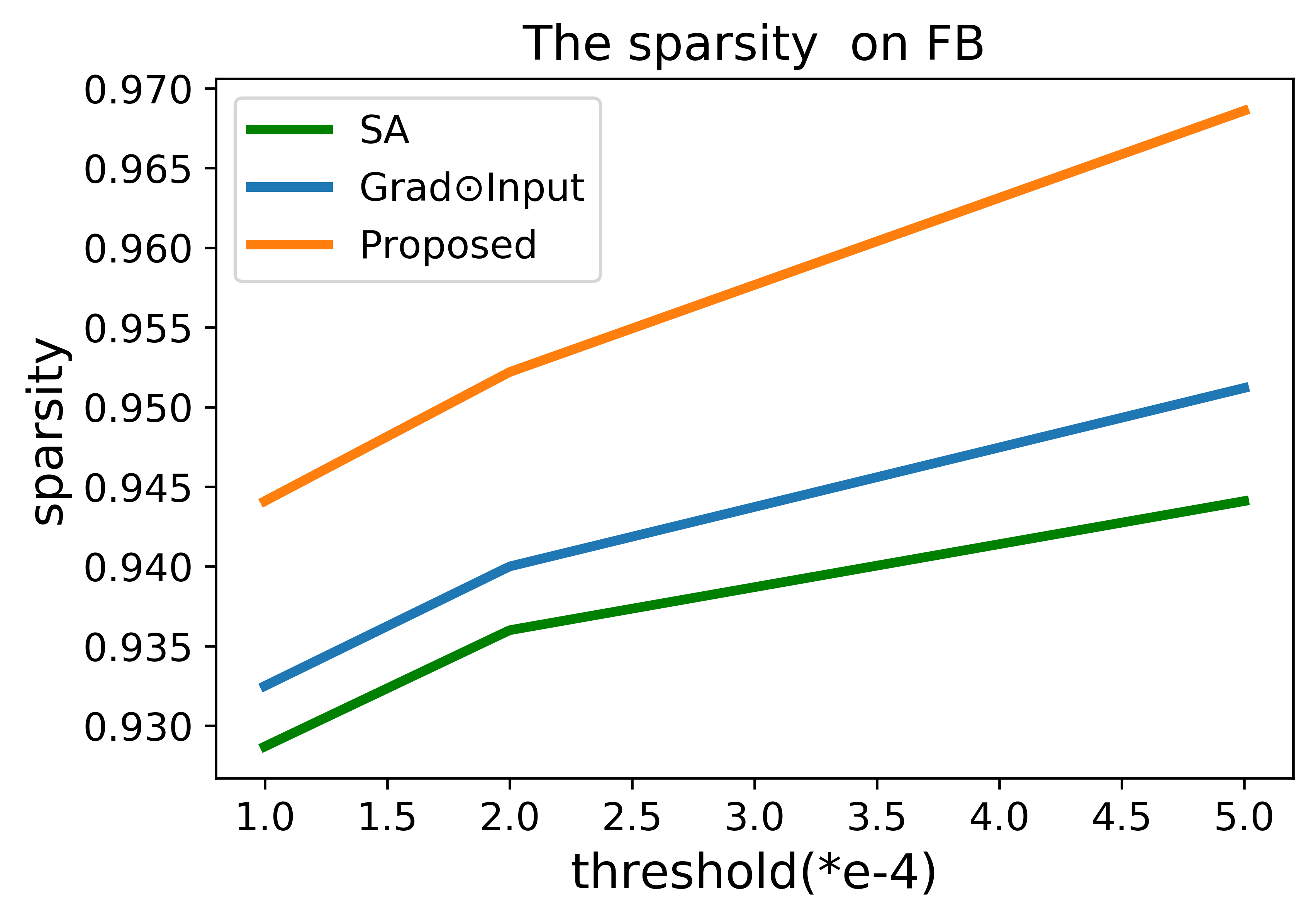}}
		\subfloat[(f) 
		COLAB]{\includegraphics[scale=0.31]{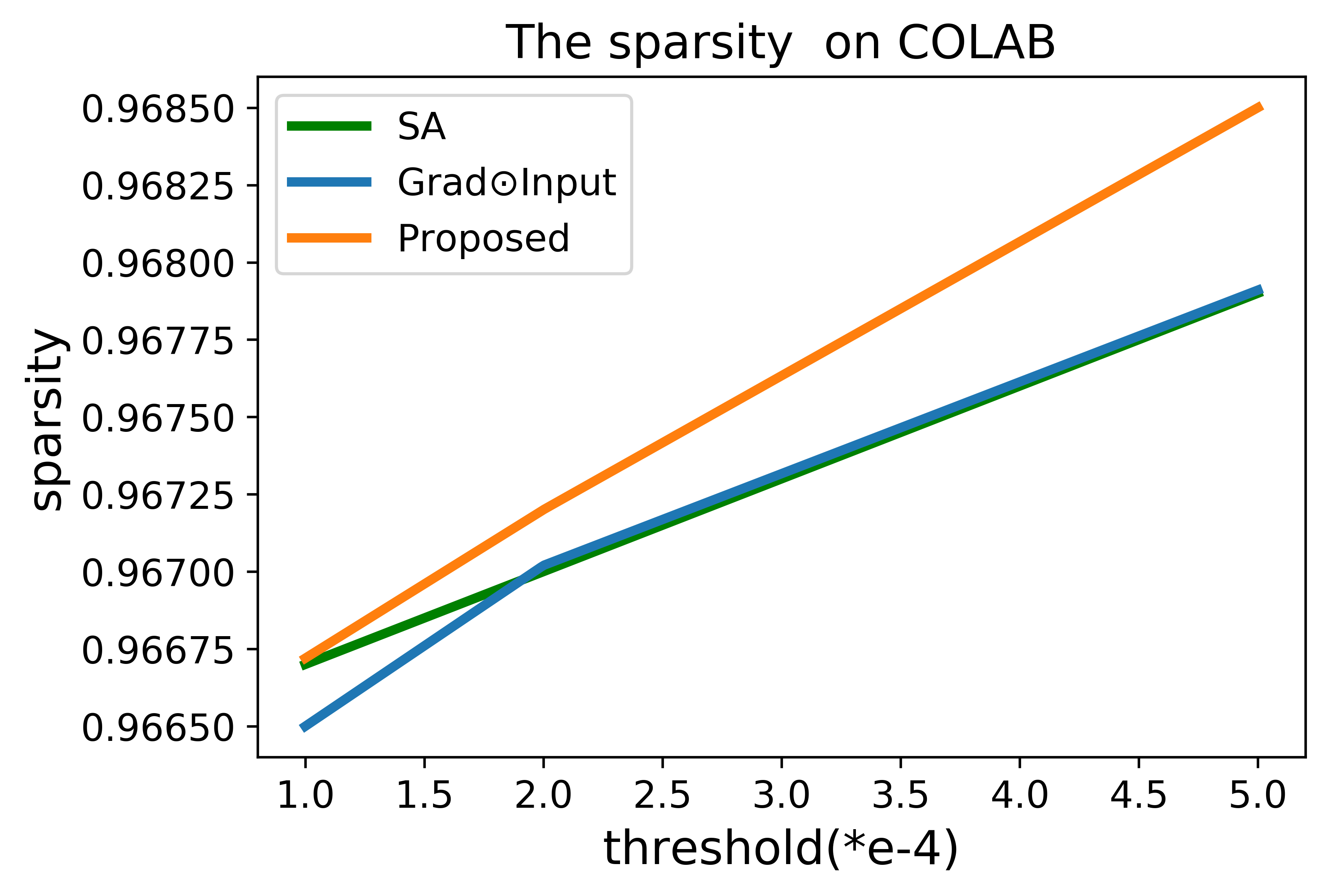}}
	\end{minipage}
	\caption{Comparison of sparsity measurement on different datasets.}
	\label{fig:sparsity}
\end{figure*}
\subsubsection{Evaluation indexes}
Next, we also report quantitative metrics that capture
desirable aspects of explanations: sparsity, stability, and fidelity. Following these metrics, we compare the quality of each explanation baseline and demonstrate the trade-offs.
\begin{itemize}[leftmargin=*]
\item \textbf{Sparsity}: 
Sparsity measures the fraction of nodes that are selected for an explanation~\cite{yuan2021explainability,pope2019explainability}. It evaluates whether the model could efficiently mark the most contributive part in the dataset. High Sparsity scores mean that fewer numbers of noes are identified as important. In our experiment, we compute sparsity by calculating the ratio of nodes with saliency values or relevances lower than a certain pre-defined threshold on a scale of $0$ to $1$.
\item \textbf{Fidelity}:
Fidelity is used to characterize whether the explanations are faithfully important to the model predictions~\cite{2020attributes}. In the experiment, we obtain fidelity by calculating the difference in classification accuracy or regression errors obtained by occluding all nodes with importance values greater than a threshold, on a scale of $0$ to $1$. And we averaged the fidelity scores across classes for each method.

\item \textbf{Stability}:
Stability measures the explanation when small changes are applied to the input~\cite{2020attributes}. Good explanations should be stable, which means that the explanations remain the same when there are small changes. 
To evaluate the stability of methods under input perturbations, we add $20\%$ more edges randomly and evaluate the change of relevances/importances produced by the model.
\vspace{-3mm}
\subsection{Experimental settings}
We run all of our experiments on an NVIDIA RTX A4000 Ti GPU with 16GB of RAM. For all datasets, we consider a two-layer GCN. We train the model on the training set for $100$ epochs with the Adam optimizer \cite{kingma2014adam} and with the initial learning rate of $0.01$. When necessary, we employ a two-layer MLP with $64$ hidden units. For the proposed model, we fix the $\epsilon=0.0001$. Note that the model performance results are obtained on average analysis. Our evaluation metrics include \textit{sparsity, fidelity} and \textit{stability}. 
\end{itemize}
\subsection{Quantitative Evaluation}
In this section, we show the effectiveness of our framework in capturing those important nodes 
in terms of fidelity, sparsity, and stability.
\hfill\\
\noindent \textbf{Results on Fidelity.}
Fidelity reflects whether an explanation method is capable of capturing important nodes.
A good explanation method should have high fidelity. To measure fidelity, we first sort the nodes from high to low according to their importance, and then we occlude several top nodes and keep other nodes reserved (\textit{e.g.} remained 90\%, 80\%, 70\%, 60\%, 50\%). 
As illustrated in Fig. (\ref{fig:fidelity}),  the proposed method achieves the best performance on most datasets and obtains comparable results on the remaining datasets.
\hfill\\
\noindent \textbf{Results on Sparsity.}
According to different thresholds, we plot the sparsity measurements as illustrated in Fig. (\ref{fig:sparsity}). We can see the proposed method achieves the highest sparsity.
\hfill\\
\noindent \textbf{Results on Stability.}
A stable explanation method should give close explanations when the inputs have small perturbations. Hence, for stability indexes, the lower, the better. As illustrated in Table. (\ref{tab:stability}), the proposed method is better than Grad$\odot$Input but not as satisfactory as SA.  
		\begin{table}[!htp]
		\small
			\setlength{\tabcolsep}{4pt}
	\centering
	\caption{The stability performance on the datasets.}
	
	\begin{tabular}{lllllll}
		\toprule
		Dataset  &   RH & PeMS04  & PeMS08  & Enron   & FB   & COLAB \\ 
		\midrule
		SA&   0.172 &0.183 & 0.151 &	0.130  &    0.130   &0.177 \\
		Grad$\odot$Input& 0.286 &0.221 & 0.157& 0.148&0.151& 0.245 \\
	     DGExplainer&0.211 &0.201 & 0.160 & 0.146       &       0.149         &                  0.248  \\
		\bottomrule
	\end{tabular}
	\label{tab:stability}
\end{table}
\subsection{Qualitative Results}
\begin{figure}[H]
	\centering
	\includegraphics[width=0.8\linewidth]{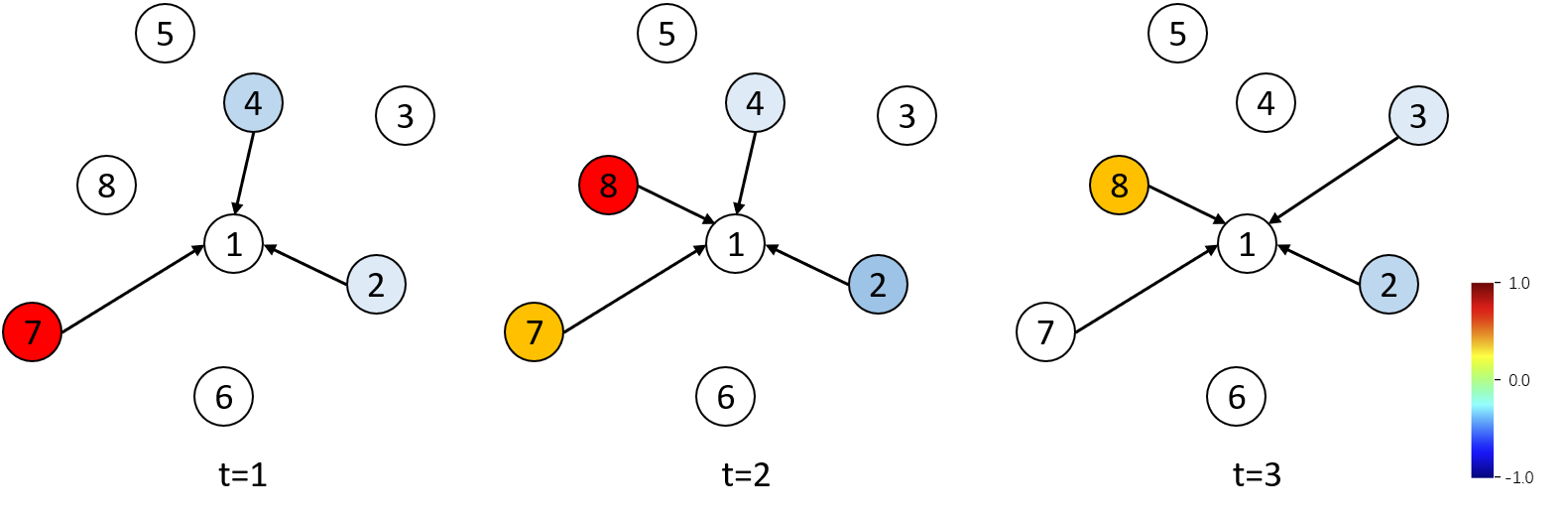}
	\caption{The illustration of the proposed method on PeMS04 dataset. The warm color shows positive effects and the cold color shows negative effects. The depth of the color reflexes the degree of the effect.
	} 
	\label{fig:pems04}
	\vspace{-3mm}
\end{figure}

In addition to the quantitative experiments, we also visualize the explanation of PeMS04 and demonstrate the effectiveness of our proposed framework in Fig. (\ref{fig:pems04}).
In this example, we show three graphs from time stamp $t=1$ to $t=3$ from PeMS04 dataset. 
Our model successfully identifies node $7$ has the largest positive effect on the prediction. And at $t=2$, node $7$ begins to have traffic to node $1$, and node $8$ is identified as the most contributive node that positively affects the prediction at this time.
\vspace{-3mm}
\section{Conclusion}\label{sec:conclusion}
In this paper, we study the explainability of DGNNs and propose a practical framework, named DGExplainer, that relies on the backward relevance to yield explanations for the prediction of DGNNs. We systematically evaluate the performance of representative explainability methods for DGNNs using the metrics of \textit{sparsity, fidelity} and stability, meanwhile conducting qualitative experiments to illustrate the effectiveness of the proposed model. The results demonstrate that DGExplainer has a satisfactory performance. We hope this framework can benefit the machine learning community in dynamic graph neural networks.
\bibliographystyle{IEEEtran}
\bibliography{refs.bib}

\begin{thebibliography}{10}
\providecommand{\url}[1]{#1}
\csname url@samestyle\endcsname
\providecommand{\newblock}{\relax}
\providecommand{\bibinfo}[2]{#2}
\providecommand{\BIBentrySTDinterwordspacing}{\spaceskip=0pt\relax}
\providecommand{\BIBentryALTinterwordstretchfactor}{4}
\providecommand{\BIBentryALTinterwordspacing}{\spaceskip=\fontdimen2\font plus
\BIBentryALTinterwordstretchfactor\fontdimen3\font minus
  \fontdimen4\font\relax}
\providecommand{\BIBforeignlanguage}[2]{{%
\expandafter\ifx\csname l@#1\endcsname\relax
\typeout{** WARNING: IEEEtran.bst: No hyphenation pattern has been}%
\typeout{** loaded for the language `#1'. Using the pattern for}%
\typeout{** the default language instead.}%
\else
\language=\csname l@#1\endcsname
\fi
#2}}
\providecommand{\BIBdecl}{\relax}
\BIBdecl

\bibitem{kipf2016semi}
T.~N. Kipf and M.~Welling, ``Semi-supervised classification with graph
  convolutional networks,'' \emph{arXiv preprint arXiv:1609.02907}, 2016.

\bibitem{hamilton2017inductive}
W.~L. Hamilton, R.~Ying, and J.~Leskovec, ``Inductive representation learning
  on large graphs,'' \emph{arXiv preprint arXiv:1706.02216}, 2017.

\bibitem{sankar2018dynamic}
A.~Sankar, Y.~Wu, L.~Gou, W.~Zhang, and H.~Yang, ``Dynamic graph representation
  learning via self-attention networks,'' \emph{arXiv preprint
  arXiv:1812.09430}, 2018.

\bibitem{ma2020streaming}
Y.~Ma, Z.~Guo, Z.~Ren, J.~Tang, and D.~Yin, ``Streaming graph neural
  networks,'' in \emph{SIGIR}, 2020, pp. 719--728.

\bibitem{pareja2020evolvegcn}
A.~Pareja, G.~Domeniconi, J.~Chen, T.~Ma, T.~Suzumura, H.~Kanezashi, T.~Kaler,
  T.~Schardl, and C.~Leiserson, ``Evolvegcn: Evolving graph convolutional
  networks for dynamic graphs,'' in \emph{AAAI}, 2020, pp. 5363--5370.

\bibitem{huang2020graphlime}
Q.~Huang, M.~Yamada, Y.~Tian, D.~Singh, D.~Yin, and Y.~Chang, ``Graphlime:
  Local interpretable model explanations for graph neural networks,''
  \emph{arXiv preprint arXiv:2001.06216}, 2020.

\bibitem{baldassarre2019explainability}
F.~Baldassarre and H.~Azizpour, ``Explainability techniques for graph
  convolutional networks,'' in \emph{ICML Workshops}, 2019.

\bibitem{pope2019explainability}
P.~E. Pope, S.~Kolouri, M.~Rostami, C.~E. Martin, and H.~Hoffmann,
  ``Explainability methods for graph convolutional neural networks,'' in
  \emph{CVPR}, 2019, pp. 10\,772--10\,781.

\bibitem{rudin2019stop}
C.~Rudin, ``Stop explaining black box machine learning models for high stakes
  decisions and use interpretable models instead,'' \emph{Nature Machine
  Intelligence}, vol.~1, no.~5, pp. 206--215, 2019.

\bibitem{ying2019gnnexplainer}
R.~Ying, D.~Bourgeois, J.~You, M.~Zitnik, and J.~Leskovec, ``G{NNE}xplainer:
  Generating explanations for graph neural networks,'' in \emph{NeurIPS},
  vol.~32, 2019, p. 9240.

\bibitem{luo2020parameterized}
D.~Luo, W.~Cheng, D.~Xu, W.~Yu, B.~Zong, H.~Chen, and X.~Zhang, ``Parameterized
  explainer for graph neural network,'' \emph{arXiv preprint arXiv:2011.04573},
  2020.

\bibitem{schlichtkrull2020interpreting}
M.~S. Schlichtkrull, N.~De~Cao, and I.~Titov, ``Interpreting graph neural
  networks for nlp with differentiable edge masking,'' \emph{arXiv preprint
  arXiv:2010.00577}, 2020.

\bibitem{chang2018explaining}
C.-H. Chang, E.~Creager, A.~Goldenberg, and D.~Duvenaud, ``Explaining image
  classifiers by counterfactual generation,'' \emph{arXiv preprint
  arXiv:1807.08024}, 2018.

\bibitem{zhang2018top}
J.~Zhang, S.~A. Bargal, Z.~Lin, J.~Brandt, X.~Shen, and S.~Sclaroff, ``Top-down
  neural attention by excitation backprop,'' \emph{IJCV}, vol. 126, no.~10, pp.
  1084--1102, 2018.

\bibitem{bach2015pixel}
S.~Bach, A.~Binder, G.~Montavon, F.~Klauschen, K.-R. M{\"u}ller, and W.~Samek,
  ``On pixel-wise explanations for non-linear classifier decisions by
  layer-wise relevance propagation,'' \emph{PloS one}, vol.~10, no.~7, 2015.

\bibitem{2020Explainability}
P.~E. Pope, S.~Kolouri, M.~Rostami, C.~E. Martin, and H.~Hoffmann,
  ``Explainability methods for graph convolutional neural networks,'' in
  \emph{CVPR}, 2019, pp. 10\,772--10\,781.

\bibitem{rebuffi2020there}
S.-A. Rebuffi, R.~Fong, X.~Ji, and A.~Vedaldi, ``There and back again:
  Revisiting backpropagation saliency methods,'' in \emph{CVPR}, 2020, pp.
  8839--8848.

\bibitem{kumar2018community}
S.~Kumar, W.~L. Hamilton, J.~Leskovec, and D.~Jurafsky, ``Community interaction
  and conflict on the web,'' in \emph{WWW}, 2018, pp. 933--943.

\bibitem{attentionbased2019}
S.~Guo, Y.~Lin, N.~Feng, C.~Song, and H.~Wan, ``Attention based
  spatial-temporal graph convolutional networks for traffic flow forecasting,''
  in \emph{AAAI}, 2019, pp. 922--929.

\bibitem{enron2004}
B.~Klimt and Y.~Yang, ``The enron corpus: A new dataset for email
  classification research,'' in \emph{ECML PKDD}.\hskip 1em plus 0.5em minus
  0.4em\relax Springer, 2004, pp. 217--226.

\bibitem{trivedi2018dyrep}
R.~Trivedi, M.~Farajtabar, P.~Biswal, and H.~Zha, ``Dyrep: Learning
  representations over dynamic graphs,'' in \emph{ICLR}, 2019.

\bibitem{rahman2016link}
M.~Rahman and M.~Al~Hasan, ``Link prediction in dynamic networks using
  graphlet,'' in \emph{ECML PKDD}.\hskip 1em plus 0.5em minus 0.4em\relax
  Springer, 2016, pp. 394--409.

\bibitem{hajiramezanali2019variational}
E.~Hajiramezanali, A.~Hasanzadeh, K.~Narayanan, N.~Duffield, M.~Zhou, and
  X.~Qian, ``Variational graph recurrent neural networks,'' in \emph{NeurIPS},
  vol.~32, 2019.

\bibitem{shrikumar2017learning}
A.~Shrikumar, P.~Greenside, and A.~Kundaje, ``Learning important features
  through propagating activation differences,'' in \emph{ICML}, 2017, pp.
  3145--3153.

\bibitem{yuan2021explainability}
H.~Yuan, H.~Yu, J.~Wang, K.~Li, and S.~Ji, ``On explainability of graph neural
  networks via subgraph explorations,'' in \emph{ICML}.\hskip 1em plus 0.5em
  minus 0.4em\relax PMLR, 2021, pp. 12\,241--12\,252.

\bibitem{2020attributes}
B.~Sanchez-Lengeling, J.~Wei, B.~Lee, E.~Reif, P.~Wang, W.~W. Qian,
  K.~McCloskey, L.~Colwell, and A.~Wiltschko, ``Evaluating attribution for
  graph neural networks,'' in \emph{NeurIPS}, vol.~33, 2020.

\bibitem{kingma2014adam}
D.~P. Kingma and J.~Ba, ``Adam: A method for stochastic optimization,''
  \emph{arXiv preprint arXiv:1412.6980}, 2014.

\end{thebibliography}
\end{document}